\documentclass[journal,twocolumn,10pt]{IEEEtran}
\usepackage{amsfonts}
\usepackage{amsmath,cases,amssymb}
\usepackage[dvips]{graphicx}
\usepackage{subfigure}
\usepackage{setspace}
\usepackage{fixltx2e}
\usepackage{cite}
\usepackage[dvipsnames]{xcolor}
\usepackage{epsfig}
\usepackage{url}
\usepackage{algorithm}
\usepackage{algorithmic}
\usepackage{multirow}
\usepackage{mathtools}
\usepackage{lipsum}
\usepackage{amsmath}
\usepackage{dblfloatfix}
\usepackage{epstopdf}
\usepackage{lineno}

\ifCLASSINFOpdf

\else
\fi

\begin{document}
%
\title{Federated Fuzzy Neural Network with Evolutionary Rule Learning}

\author{Leijie Zhang, Ye Shi$^{*}$, \textit{Member, IEEE}, Yu-Cheng Chang and Chin-Teng Lin$^{*}$, \textit{Fellow, IEEE}
\thanks{Ye Shi and Chin-Teng Lin are the corresponding authors.
}
\thanks{Leijie Zhang, Yu-Cheng Chang and Chin-Teng Lin are with CIBCI Lab, the Australian Artificial Intelligence Institute, the School of Computer Science, University of Technology, Sydney, NSW 2007, Australia. Email:
Leijie.Zhang@student.uts.edu.au, Yu-Cheng.Chang@uts.edu.au, Chin-Teng.Lin@uts.edu.au. Ye Shi is with the School of Information Science and Technology, ShanghaiTech University, Shanghai, 201210, China. Email:
shiye@shanghaitech.edu.cn.}
}

\maketitle

\begin{abstract}
Distributed fuzzy neural networks (DFNNs) have attracted increasing attention recently due to their learning abilities in handling data uncertainties in distributed scenarios. However, it is challenging for DFNNs to handle cases in which the local data are non-independent and identically distributed (non-IID). In this paper, we propose a federated fuzzy neural network (FedFNN) with evolutionary rule learning (ERL) to cope with non-IID issues as well as data uncertainties. The FedFNN maintains a global set of rules in a server and a personalized subset of these rules for each local client. ERL is inspired by the theory of biological evolution; it encourages rule variations while activating superior rules and deactivating inferior rules for local clients with non-IID data. Specifically, ERL consists of two stages in an iterative procedure: a rule cooperation stage that updates global rules by aggregating local rules based on their activation statuses and a rule evolution stage that evolves the global rules and updates the activation statuses of the local rules. This procedure improves both the generalization and personalization of the FedFNN for dealing with non-IID issues and data uncertainties. Extensive experiments conducted on a range of datasets demonstrate the superiority of the FedFNN over state-of-the-art methods. Our code is available online\footnote{\url{https://github.com/leijiezhang/FedFNN}}
\end{abstract}


\section{Introduction}

\begin{figure}[!ht]
  \centering
  \subfigure[\normalsize The evolvement of variants within a species via the selective activation of genes.]{
  \includegraphics[width=1.0\columnwidth]{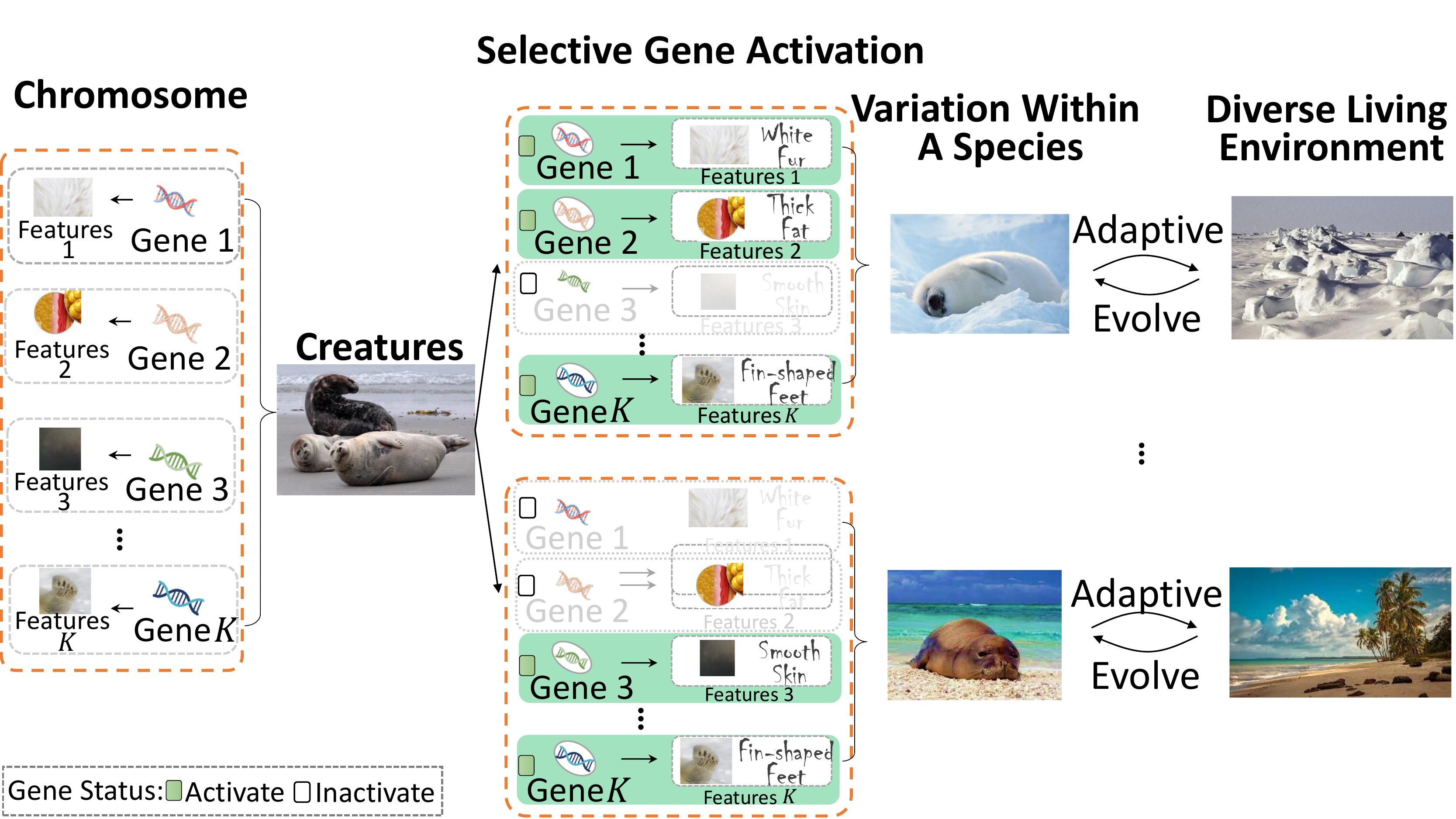}
  }
  \subfigure[\normalsize The rule selective activation for FedFNN to generate personalized local FNNs.]{
  \includegraphics[width=1.0\columnwidth]{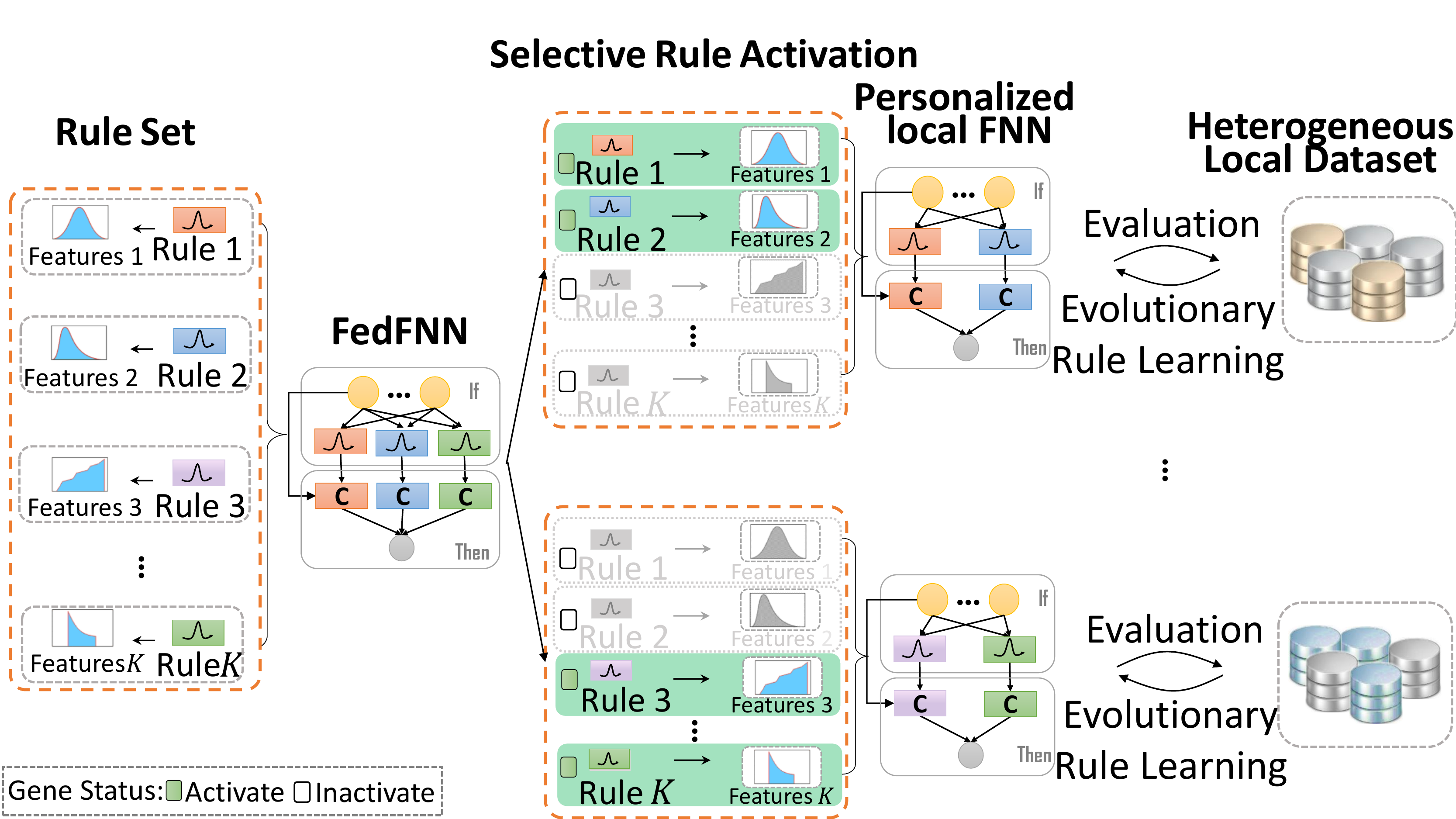}
  }

  \vspace{1pt}
  \centering
  \caption{\normalsize (a) A brief demonstration of how a species evolve variants to survive in diverse living environments based on genes selective activation and expressions. (b) A brief demonstration of how FedFNN selectively activate a personalized subset of contributive rules for clients to effectively deal with their local non-IID data.}
  \label{fig:idea}
\end{figure}
Combining neural networks with fuzzy logic \cite{hayashi1995implementation, takagi1985fuzzy}, fuzzy neural networks (FNNs) \cite{lin1991neural, jang1993anfis, ebadzadeh2017ic,buckley1994fuzzy,Shihabudheen2018} have been proposed with powerful learning capabilities and uncertainty handling abilities in centralized scenarios. However, due to increasing data privacy concerns, the samples collected from distributed parties must be processed locally. Distributed FNNs (DFNNs)\cite{fierimonte2016distributed, fierimonte2017distributed, shi2020consensus, zhang2020hierarchical, shi2021distributed} address this issue by learning a global FNN via the integration of local models. However, the existing DFNNs are fragile to non-independent and identically distributed (non-IID) data. In addition, as DFNNs tend to learn a shared group of global rules for all clients, their personalization ability is limited and their learned global rules are less adaptive. Furthermore, they regard the local model integration process as a convex optimization problem and solve it with the alternating direction method of multipliers \cite{boyd2011convex}, which overlooks the powerful learning ability of feedforward FNNs.

Fortunately, federated learning (FL) models\cite{mcmahan2017communication, bonawitz2019towards, yang2019federated} offer decentralized learning architectures that allow local models to optimize their parameters using gradient descent. The FL approach learns a shared model by aggregating the updates obtained from local clients without accessing their data. Notably, data distribution heterogeneity
is also one of the key challenges for FL. Yet, many efforts \cite{karimireddy2020scaffold,li2020federated,shamsian2021personalized, achituve2021personalized,linsner2021approaches} have been made to handle this issue in FL, in which many approaches consider solutions that allow clients to have personalized models. However, few of the existing FL methods are able to simultaneously cope with data uncertainties and non-IID issues in distributed learning scenarios. Although several studies have adopted Bayesian treatments \cite{wang2020federated} and Gaussian processes \cite{linsner2021approaches, achituve2021personalized} to enable FL methods to handle data uncertainties, their performance heavily relies on the learning of good posterior inferences and kernel functions, which is very time-consuming.

To solve the aforementioned issues, in this paper, we propose a federated fuzzy neural network (FedFNN) with evolutionary rule learning (ERL) to handle data uncertainties and non-IID issues. As shown in Fig. \ref{fig:idea} (a), the theory of biological evolution \cite{wade1990causes} states that variants of the same species can evolve to adapt to their different living environments by selectively activating and expressing their genes. Inspired by this, we use FNNs as our local models and consider them as compositions of fuzzy rules, which capture valuable local data information from multiple views, such as distributions. Similar to the genes of a species, each rule of the FedFNN is a basic functional component that can be activated or deactivated for clients according to their performance on local data. Thus, our FedFNN aims to obtain a group of global fuzzy rules that can be selectively activated for local clients to enable them to outperform competing approaches on non-IID data.
It is worth noting that our ERL is a novel algorithm different from genetic algorithms \cite{mirjalili2019genetic}. These two algorithms are inspired by the same theory but designed in different ways. The genetic algorithm treats the whole population as the evolution subject. It keeps selecting the fittest individuals in each generation to form more competitive populations until the performance gets stable. However, it needs to be re-adjusted whenever the environment is modified. Instead, the ERL considers the internal diversity of species, which means that a species normally includes several types of populations (subspecies) living in diverse environments. It regards subspecies as its evolution subjects and selectively optimises and shares gene groups among subspecies until they perform well in different environments. In our FedFNN, each agent that handles non-IID data corresponds to a sub-species. The ERL enables agents to cooperate with each other during their evolutions but preserve their personalities to adapt to diverse environments.
In general, our FedFNN aims at learning 1) a group of global rules that capture valuable information among local clients and 2) a rule activation strategy for each local client to ensure the personalization and superior performance of the FedFNN.

The ERL is an iterative learning approach with two stages: a rule cooperation stage, where the global rules are updated cooperatively by clients based on their rule activation statuses, and a rule evolution stage, where the activation statuses of local rules are adjusted according to their contributions to the performance of the FedFNN on non-IID local data. The former stage enhances the cooperation among clients for learning more representative global rules, which increases the generalizability of the FedFNN. In contrast, the latter stage fulfils the selective activation of rules that enable the local FNNs to be more adaptive and perform better on non-IID data, which improves the personalization of the FedFNN. For an explanation of the FedFNN model, refer to the diagram shown in Fig. \ref{fig:idea} (b).

The contributions of this paper are as follows,
\begin{itemize}
\item{We are the first to propose FedFNN that integrates fuzzy neural networks into a federated learning framework to handle data non-IID issues as well as data uncertainties in distributed scenarios. FedFNN is able to learn personalized fuzzy if-then rules for local clients according to their non-IID data.}

\item {Inspired by the theory of biological evolution, we design an ERL algorithm for the learning of FedFNN. ERL encourages the global rules to evolve while selectively activating superior rules and eliminating inferior ones during the training procedure. This procedure ensures the ability of generalization and personalization of FedFNN.}
\end{itemize}

\section{Related Work}
\subsection{Distributed fuzzy neural networks}
DFNNs \cite{fierimonte2016distributed, fierimonte2017distributed, shi2020consensus, shi2021distributed, zhang2020hierarchical} have been proposed to handle the uncertainties encountered in distributed applications. The authors in \cite{fierimonte2016distributed} proposed a DFNN model that randomly sets the parameters in antecedents and only updates the parameters in consequent layers. Later, they extended this work to an online DFNN model \cite{fierimonte2017distributed}. Their models assume that all clients share the information in antecedent layers, making this technically not a seriously distributed method. To avoid this problem, a fully DFNN \cite{shi2020consensus} model was proposed by adopting consensus learning in both the antecedent and consequent layers. As its subsequence variant, a semisupervised DFNN model \cite{shi2021distributed} was presented to enable the DFNN to leverage unlabeled samples by using the fuzzy C-means method and distributed interpolation-based consistency regularization. However, the existing DFNNs cannot handle situations in which the data distribution varies across clients. The authors in \cite{zhang2020hierarchical} proposed a DFNN with hierarchical structures to process the heterogeneity existing in the variables of training samples. However, instead of processing the data heterogeneity across distributed clients, they focused on variable composition heterogeneity, which meant that data variables were collected from different sources.
Generally, by employing the well-known Takagi-Sugeno (T-S) fuzzy if-then rules \cite{takagi1985fuzzy}, the existing DFNN models build the antecedent layers of their local models in traditional ways (e.g., K-means) and calculate the corresponding consequent layers with closed-form solutions. Then, the original DFNNs are transformed into convex optimization problems. While efficient and effective, they are not able to learn local models with personalized rule sets. Worse, they fail to utilize the strong learning abilities of neural networks that enable local FNNs to investigate more adaptive rules.

\subsection{Federated Learning}
FL \cite{mcmahan2017communication} is an emerging distributed paradigm in which multiple clients cooperatively train a neural network without revealing their local data. Recently, many solutions\cite{mcmahan2017communication, mohri2019agnostic, singh2020model} have been presented to solve FL problems, among which the most known and basic solution is federated averaging (FedAvg) \cite{mcmahan2017communication}, which aggregates local models by calculating the weighted average of their updated parameters.

However, FL has encountered various challenges \cite{zhang2021survey}, among which the non-IID issue is the core problem that makes the local model aggregation process harder and leads to performance degradation. Numerous FL algorithms have been presented to solve the non-IID problem, e.g., stochastic controlled averaging for FL (SCAFFOLD) \cite{karimireddy2020scaffold}; FedProx \cite{li2020federated}; model-contrasted FL (MOON) \cite{li2021model}, which attempts to increase the effect of local training on heterogeneous data by minimizing the dissimilarity between the global model and local models; FedMA \cite{wang2020federated} and FedNova \cite{wang2020tackling}, which improve the aggregation stage by utilizing Bayesian nonparametric methods and local update normalization, respectively; CCVR \cite{luo2021no}, which calibrates the constitutive classifiers using virtual representations to eliminate the global model bias caused by local distribution variance; and FedEM \cite{dieuleveut2021federated}, which introduces expectation maximization to make the learned model robust to data heterogeneity. Though these methods have been proposed based on FedAvg by trying to learn a more robust global model, they focus on learning a shared global model, which degrades their performance when the data distributions heavily vary across clients.

Recently, personalized FL (PFL) \cite{kulkarni2020survey, achituve2021personalized,shamsian2021personalized, fallah2020personalized} has been proposed; this approach aims to process heterogeneous local data with personalized models. Many of the existing PFL methods were proposed to solve the distributed meta-learning problem \cite{fallah2020convergence, fallah2020personalized, huang2021personalized, tang2021distributed}. Among the methods that target normal PL tasks, multitask learning \cite{smith2017federated} is applied to learn personalized local models by treating each client as a learning task; model mixing \cite{ma2022layer, collins2021exploiting} achieves the same goal by allowing clients to learn a mixture of the global model and local models. Notably, the authors in \cite{singhal2021federated} presented a new local model structure that comprises a global feature encoder and a personalized output layer. By contrast, LG-FedAvg provides clients with a local feature encoder and a global output layer.

However, very few of the mentioned FL methods are able to handle data uncertainties, except for that of \cite{wang2020federated, wang2020tackling}, which adopts Bayesian treatment, and that of \cite{achituve2021personalized}, which adopts a Gaussian process. In addition, building Bayesian posteriors and Gaussian kernels is time-consuming. In contrast, our study uses FNNs as local models, which are viewed as assemblies of fuzzy rules. Thus, taking rules as basic functional units, we break down the task of learning a global model into learning global fuzzy rules, each of which can independently investigate its local sample space and contribute to the local training process.

\section{Federated Fuzzy Neural Network}
\begin{figure*}[!htbp]
  \centering
  \includegraphics[width=1.2\columnwidth]{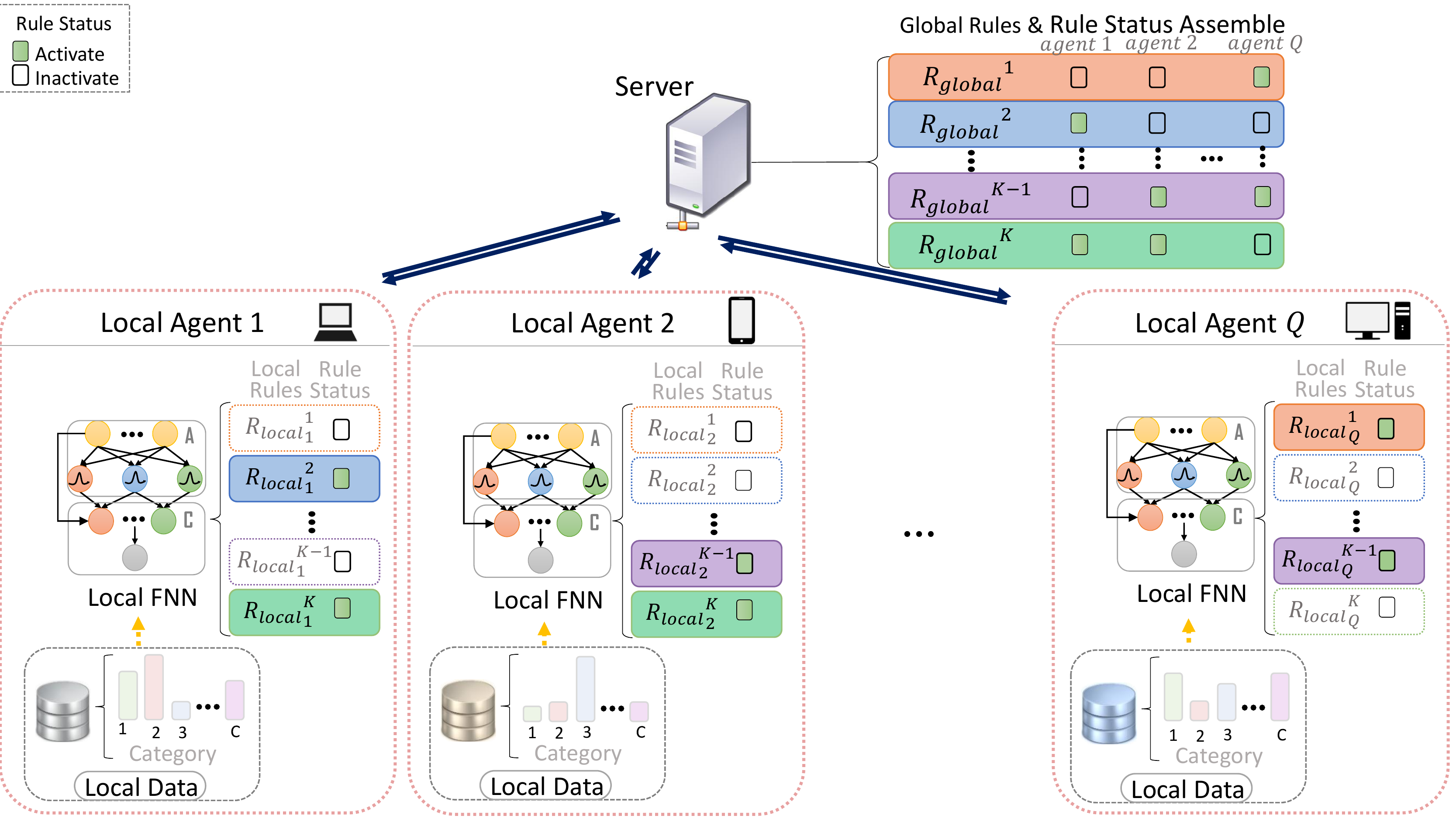}
  \vspace{-10pt}
  \caption{\normalsize Overview of the FedFNN.}
  \label{fig:fed_fnn}
\end{figure*}

\begin{figure*}[!htbp]
  \centering
  \includegraphics[width=1.2\columnwidth]{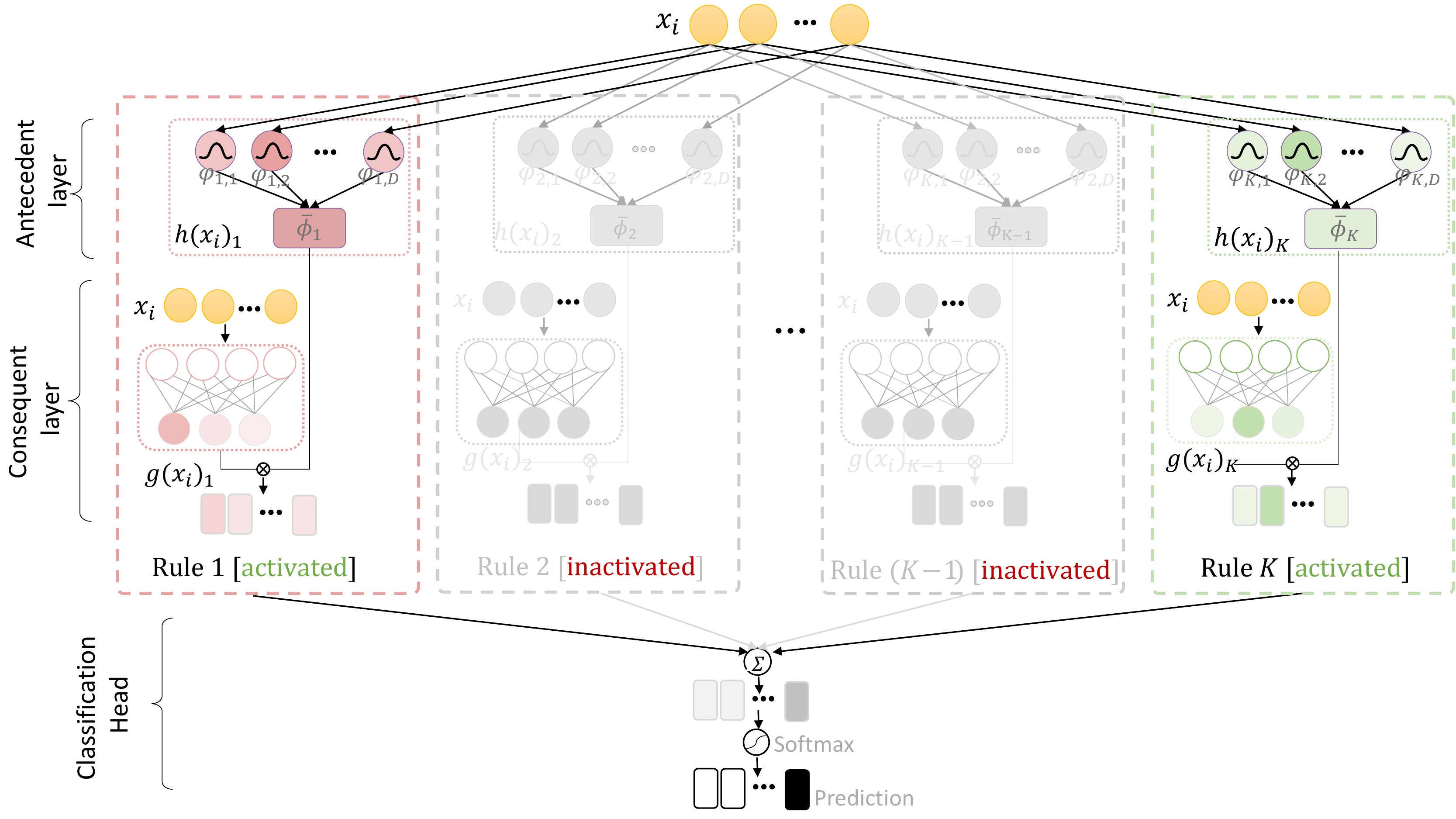}
  \vspace{-10pt}
  \caption{\normalsize Overview of a local FNN.}
  \label{fig:local_fnn}
\end{figure*}

In this section, we describe the general structure of our FedFNN. As depicted in Fig. \ref{fig:fed_fnn}, the FedFNN includes one server and several local clients. The server is responsible for communication with local clients and maintaining a group of global rules by aggregating the uploaded local rules. Local clients download the global rules as local rules for constructing their FNNs, which are then updated via training on their own data. Due to the concerns of data privacy, each local agent learns from its own data without accessing the data of other agents and communicates, while the server communicates with all local agents and aggregates the locally learned rules according to their activation status.
An overview of a local FNN is shown in Fig. \ref{fig:local_fnn}.

To mimic the selective activation of genes, the rules in the local clients are activated selectively to make the FedFNN personalized and properly adapted to local non-IID data. Thus, we introduce an activation vector containing the rule status $s_k^q$ of each client. Accordingly, the global server can help local clients activate useful rules and deactivate useless or harmful rules based on their own local data. For example, if $s_k^q = 0$, the server will deactivate the $k$-th rule for the FNN on the $q$-th client; otherwise, the corresponding rule will be activated and involved in the operations of the $q$-th client.

In the FedFNN, we adopt the fuzzy logic presented in the first-order T-S fuzzy system \cite{takagi1985fuzzy}.
Suppose that our FedFNN holds $Q$ clients; then, the dataset owned by the $q$-th client can be denoted as $\mathcal{D}^q := \{x_i^q,y_i^q\}_{i=1}^{N^q}$, where $x_i^q = [x_{i1}^q, x_{i2}^q,\cdots,x_{iD}^q]^T$ and $y_i^q\in \mathbb{R}^C$ are the $i$-th sample and its one-hot vector label, respectively, and $C$ and ${N^q}$ denote the category number and the local dataset size.
Then, the $k$-th fuzzy rule of the local FNN on the $q$-th client can be described as
\begin{center}
  Rule $k$: If $x_{i1}^q$ is $A_{k1}^q$ and $\cdots$ and $x_{iD}^q$ is $A_{kD}^q$\\
  Then, $y_i^q$ = ${g^q(x_i;\theta_k)}$
\end{center}
where $A_{kj}^q$ is the $j$-th fuzzy set of rule $k$ on $P^q$ and $g^q(x_i;\theta_k)$ is the corresponding consequent rule built by a fully connected layer parameterized with $\theta_k$. Many types of membership functions can be employed to describe the fuzzy set $A_{kj}^q$, such as singleton, triangular, trapezoidal, and Gaussian ones \cite{jang1997neuro}. Here we choose the Gaussian membership function for three reasons: 1) Gaussian membership function is differentiable, which is more suitable for end-to-end learning models; 2) Gaussian membership function is proved to be effective in approximating nonlinear functions on a compact set \cite{juang1998online}; 3) Gaussian membership functions are able to represent data features using different Gaussian distributions, which can capture data heterogeneity and handle data uncertainty using several distributions.

Generally, the Gaussian membership function $\varphi^q(x_{ij}^q;m_{kj}^q, \sigma_{kj}^q)$ of a fuzzy set $A_{kj}^q$ has the form
\begin{equation}\label{Gaussian_A}
\varphi_k^q(x_{ij}^q;m_{kj}^q, \sigma_{kj}^q) = \text{exp}{\left[-\left(\frac{x_{ij}^q-m_{kj}^q}{\sigma_{kj}^q}\right)^2\right]},
\end{equation}
where $m_{kj}$ and $\sigma_{kj}$ are the mean and standard deviation of the Gaussian membership function, respectively. Let $\varphi_{k}^q(x_i^q;m_k^q, \sigma_k^q)$ collect the membership value of the $k$-th rule on client $P^q$ in vector form. The output of the antecedent layer (also known as the firing strength) considering the rule activation status $s_k^q$ can be written as:
\begin{equation}\label{func_antece_s}
{h_k^q(x_i^q;m_k^q,\sigma_k^q,s_k^q)} =  \frac{s_k^q\text{exp}{[||\varphi_{k}^q(x_i^q;m_k^q, \sigma_k^q)||_2]}}{\sum_{k=1}^{K}s_k^q\text{exp}{[||\varphi_k^q(x_i^q;m_k^q, \sigma_k^q)||_2]}},
\end{equation}
The consequent output of the $k$-th rule is denoted as ${g^q(x_i^q;\theta_k^q)}$ and can be calculated by:
\begin{equation}\label{func_consq}
{g_k^q(x_i^q;\theta_{k})} = [1;x_i^q]^T\theta_k^q,
\end{equation}
where $\theta_k^q \in \mathbb{R}^{(D+1)\times C}$ denotes the consequent parameters.
Thus, by considering rule statuses, the FedFNN is able to eliminate the interruption of deactivated rules for local clients. The antecedent layer $h^q(x_i^q;m^q,\sigma^q)$ and the consequent layer $g^q(x_i^q;\theta^q)$ of the $q$-th client in our model are
\begin{equation}\label{layer_antece}
{h^q(x_i^q;m^q,\sigma^q)} = (h_1^q(x_i^q;m_1,\sigma_1,s_1^q),\cdots,h_K^q(x_i;m_k^q,\sigma_k^q,s_K^q)),
\end{equation}
and
\begin{equation}\label{layer_consq}
{g^q(x_i^q;\theta^q)} = (g_1^q(x_i^q;\theta_1),\cdots,g_K^q(x_i^q;\theta_k^q)),
\end{equation}
respectively. A classification head is further added to the tail to generate the final predictions by considering the outputs of all $K$ rules. The classification head connects all rules, which guarantees that the gradient of the loss function can successfully propagate backward to every component of the local FNNs.
Thus, the predictions of the $q$-th local FNN $f^q(x_i^q;m^q,\sigma^q,\theta^q,s^q)$ can be represented as:
\begin{equation}\label{model_local_fnn}
{f^q(x_i^q;m^q,\sigma^q,\theta^q,s^q)} = \text{softmax}\left(\tau\right),
\end{equation}
where $\tau = \sum_{k=1}^{K}{h_k^q(x_i^q;m_k^q,\sigma_k^q,s_k^q)}{g_k^q(x_i^q;\theta_k^q)}$.
Suppose that $w^q=(m^q,\sigma^q,\theta^q)$ collect the parameters of all local rules on client $q$; the loss of the $q$-th local FNN can then be defined as
\begin{equation}\label{loss_local_fnn}
{\ell^q(x_i^q;w^q,s^q)} = -\sum_{c=1}^{C}{y_{ic} \log(\hat{y}_{ic})},
\end{equation}
where $\hat{y}_{ic}$ is the $c$-th output of ${f^q(x_i^q;w^q,s^q)}$,
and its goal is to optimize
\begin{equation}\label{obj_local_fnn}
\min_{w^q, s^q}\mathbb{E}_{(x,y)\sim \mathcal{D}^q}[\ell^q((x,y);w^q, s^q)].
\end{equation}

Thus, the training objective of our proposed FedFNN can be given by
\begin{equation}\label{obj_fed_fnn}
\begin{aligned}
\arg \min_{\Theta}\frac{1}{Q}&\sum_{q=1}^{Q}\ell^q(x_i^q;w^q,s^q)\\ &=\arg \min_{\Theta}\frac{1}{Q}\sum_{q=1}^{Q}\frac{1}{N^q}\sum_{i=1}^{N^q}\ell^q(x_i^q;w^q,s^q),
\end{aligned}
\end{equation}
where $\Theta$ denotes the set of personal parameters $\{w^q, s^q\}_{q=1}^Q$.

\section{Evolutionary Rule Learning}
\begin{figure*}[!htbp]
  \centering
  \includegraphics[width=1.2\columnwidth]{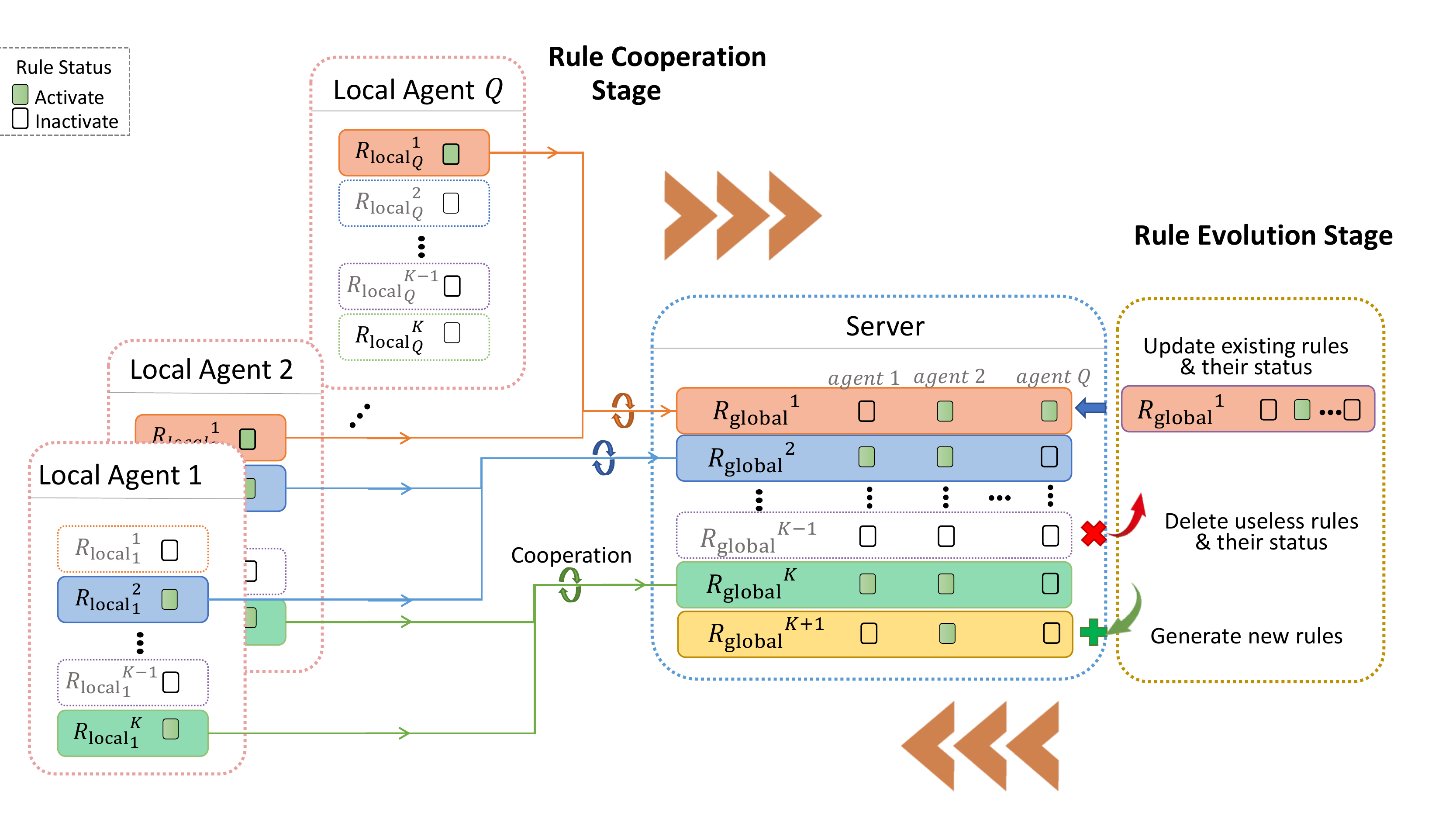}
  \vspace{-10pt}
  \caption{\normalsize Overview of the evolutionary rule learning.}
  \label{fig:learn_stg}
\end{figure*}

In this section, we present an in-depth introduction of the ERL method, which enables the FedFNN to be personalized and achieve superior performance on non-IID data.
As shown in Fig. \ref{fig:learn_stg}, ERL includes a rule cooperation stage that improves the generalization of global rules and a rule evolution stage that enhances the personalization of the FedFNN. The above two stages are presented in subsection IV.A and subsection IV.B, respectively.

\begin{algorithm}
\caption{Co-Evolutionary Stage of the ERL Method} \label{alg:co}
\begin{algorithmic}
\STATE{\textbf{Input:} Number of clients $Q$, number of global rules $K$, number of local epochs $E$, learning rate $\eta$, global ruleset parameter $w(t)$.}

\STATE{\textbf{Output:} The global ruleset parameter $w(t+1)$ for the next round.}
\STATE{}
\STATE{\textbf{Server executes:}}\\
\FOR{$q=1,2,\cdots,Q$ in parallel, do}
\STATE{send global ruleset parameter $w(t)$ to $P^q$}
\STATE{send rule activation status $s^q(t)$ to $P^q$}
\STATE{receive ${w^q}(t)$ from \textbf{Local Training}}
\ENDFOR
\FOR{$k=1,2,\cdots,K$}
\STATE{\textbf{update the $k$-th rule $w_k(t+1)$} using (\ref{func_param_agg})}
\ENDFOR
\STATE{}
\STATE{\textbf{Local Agent Training:}}
\STATE{${w^q}(t) \leftarrow$ $w(t)$}
\FOR{epoch $i=0,1,2,\cdots,E$}
\FOR{each batch $\mathbf{b} = \{x,y\}$ of $\mathcal{D}^q$ }
\STATE{\textbf{calculate $\ell$} via (\ref{loss_local_fnn})}
\STATE{\textbf{update $w^q(t)$} via ${w^q}(t) \leftarrow$ ${w^q}(t) - \eta\nabla\ell$ }

\ENDFOR
\ENDFOR\\
return ${w^q}(t)$ to server

\end{algorithmic}
\end{algorithm}

\begin{algorithm}
\caption{Adaptive Evolutionary Stage of the ERL Method} \label{alg:adaptive}
\begin{algorithmic}
\STATE{\textbf{Input:} Number of clients $Q$, number of global rules $K$, global rule parameter set $w(t)$, hyperparameter $\beta$.}

\STATE{\textbf{Output:} The global ruleset parameter $w(t+1)$, the rule status $s(t+1)$, and the rule number $K$ for the next round.}
\STATE{}
\STATE{\textbf{Server executes:}}\\

\FOR{$q=1,2,\cdots,Q$ in parallel, do}

\FOR{$k=1,2,\cdots,K$ in parallel, do}
\STATE{\textbf{update ${s_k^q}(t+1)$} via (\ref{func_s_update}))}.
\ENDFOR
\IF{(\ref{rule_gene_factor1}) and (\ref{rule_gene_factor2}) both hold or $\sum{s^q}(t+1) = 0$}
    \STATE generate a new rule for client $P^q$ and randomly initialize its parameter $w_{K+1}$.
    \STATE expand global parameter set $w(t+1)$ via $w(t+1) \leftarrow$ $[w(t), w_{K+1}]$.
    \STATE expand rule status $s(t+1)$ via $s(t+1) \leftarrow$ $[s(t), \mathbf{0}]$, ${s^q_{K+1}}(t) = 1$.
    \STATE update $K$ via $K = K + 1$
\ENDIF

\ENDFOR

\FOR{$k=1,2,\cdots,K$}
\IF{$\sum_{1}^{Q}s_k^q = 0$}
\STATE remove rule $k$ from global rule list and $w_k$ from global parameter set $w$.
\STATE update $K$ via $K = K - 1$
\ENDIF
\ENDFOR
\end{algorithmic}
\end{algorithm}

\subsection{Rule Cooperation Stage}
In this stage, we focus on the learning of more general global rules. Generally, the global rules activated by multiple clients are able to capture informative representations across these clients. Thus, updating the general global rules requires cooperation among the local clients.

Technically, the rule cooperation stage highly relies on the rule activation status vectors of the local clients. These vectors are randomly initialized at the beginning and updated in the rule evolution stage. During each communication round, the local clients download the global rules from the server as their local rules. Then, the global server selects the activated rules according to the corresponding rule activation status vectors to build the local FNNs.
As described in the Local Training of the algorithm \ref{alg:co}, the constructed personalized local FNNs are then trained on their associated non-IID local data to update the activated local rules.

Afterwards, each global rule is updated by calculating the weighted average of the activated local rules.
Thus, the parameters of the $k$-th global rule $w_k$ are updated by activation-status driven weight averaging:
\begin{equation}\label{func_param_agg}
{w_k} = \sum_{q=1}^{Q}\frac{N^qs_k^q}{\gamma_k}(w_q^k),
\end{equation}
where $\gamma_k = \sum_{q=1}^{Q}N^qs_k^q$ is the factor that measures the contribution of a global rule.
This stage is conducted for $L$ rounds before stepping to the rule evolution stage to ensure that the global rules are effectively learned under cooperation among the local clients.

It is worth noting that each global rule is updated by aggregating its corresponding activated local rules instead of aggregating all of them. This is different from existing FL methods that aggregate local models without checking whether their updates are helpful or not. Our ERL approach only shares the useful pieces of information that contribute to each other while avoiding misleading information. This procedure enables the FedFNN to be more general.

\subsection{Rule Evolution Stage}
After $L$ rounds of coevolutionary learning, the FedFNN optimization process falls into a bottleneck since the capability of the current model structure has been fully explored. In this stage, we allow the server to inspect the performance of the learned model on local samples and then implement the FedFNN to increase its performance and personalization for handling non-IID data. In general, as described in algorithm \ref{alg:adaptive}, the rule evolution stage includes 3 major mutations: evolving new global rules, activating superior rules, and deactivating local inferior rules.

We adopt a contribution factor $\pi_k^q$ to measure the importance of the $k$-th rule on the $q$-th client, which can be calculated as
\begin{equation}\label{contr_lvl}
\pi_k^q = \sum_{i=1}^{|\mathcal{D}^q|}\frac{{h_k^q(x_i^q;m_k^q,\sigma_k^q,s_k^q)}}{|\mathcal{D}^q|},
\end{equation}
where $|\mathcal{D}^q|$ is the size of $\mathcal{D}^q$. Here, we introduce a threshold $\bar{\pi}^q$ to trigger the rule activation procedure. $\bar{\pi}^q$ is calculated based on the average contribution of the activated rules as $\bar{\pi}^q =  \beta\sum_{k=1}^{K}s_k^q\pi_k^q/K^q$, where $\beta$ is a hyperparameter and $K^q$ is the number of activated rules in the $q$-th client. A rule is meaningless to client $k$ when its contribution level is smaller than $\bar{\pi}$. Consequently, the server deactivates this rule for the corresponding clients and eliminates the involvement of those clients in the rule aggregation process. To assure that this step is completed, the rule status $s_k^q(t)$ in the $t$-th round is updated by a status adjusting operation:
\begin{equation}
\label{func_s_update}
{s_k^q}(t) = \left\{
\begin{array}{cc}
    1, & \pi_k^q > \bar{\pi}_k^q \\
    0, & \text{otherwise}
\end{array}
\right.
\end{equation}
It is worth noting that each client should inspect its activated rules to check if the combination of these rules is sufficient for evaluating its local dataset. If not, it means that some of the unique features are not captured by the current activated rule settings. To solve this problem, we introduce two conditions to evaluate the capabilities of local models from different views:
\begin{equation}\label{rule_gene_factor1}
\sum_{l=1}^{L}\frac{{\ell^q}(t-l+1)-{\ell^q}(t-l)}{L} > 0,
\end{equation}
\begin{equation}\label{rule_gene_factor2}
\left(\ell^q(t) - \sum_{q=1}^{Q}\frac{{\ell^q}(t)}{Q} \right) > 0,
\end{equation}
where $t$ is the current training round. The first condition (\ref{rule_gene_factor1}) serves as a self-evaluation of the learning performance for each client by monitoring the loss value trends.
If condition (\ref{rule_gene_factor1}) holds, then the current local rules cannot be further improved.
The second condition (\ref{rule_gene_factor2}) serves as a peer evaluation by comparing the loss of the $q$-th client with the average loss across all clients.
If condition (\ref{rule_gene_factor2}) holds,
then the current local rules cannot handle non-IID data well. If both of these conditions hold, then the current architecture in the $q$-th client is unable to attain high performance. In this case, a new rule should be generated to improve the local model. To avoid the extreme situation when certain agents have no rules to use, the server will also create new rules for agents once their activated rule number is detected as zero.

This stage imitates the selective activation process of genes by updating the activation status of each rule based on its contribution to the local clients. Consequently, ERL selects useful rules for the local clients from the well-learned global rules, increasing the personalization of the FedFNN. In addition, the updated rule activation schemes benefit the optimization process in the next round of the rule cooperation stage in turn because of the better rule selection effect.

\section{Experiments}

\begin{table}[!ht]
\centering
\caption{Dataset Information}
\begin{tabular}{c|c|c|c}
\hline
Dataset & Sample & Feature & Category \\ \hline
GSAD \cite{vergara2012chemical}    & 14,061 & 128     & 6        \\ \hline
FM \cite{gyamfi2018linear}     & 180    & 43      & 4        \\ \hline
WD \cite{cortez2009modeling}     & 4,898  & 11      & 7        \\ \hline
MGT \cite{dvovrak2007softening}    & 19,020 & 10      & 2        \\ \hline
SC \cite{bagirov2018algorithm}    & 58,000 & 9       & 7        \\ \hline
WIL \cite{rohra2017user}    & 2,000  & 7       & 4        \\ \hline
WFRN \cite{freire2009short}    & 5,456 & 24      & 4        \\ \hline
\end{tabular}
\label{tbl:dataset_detail}
\end{table}

In this section, we conduct extensive experiments on 7 datasets of different types in various settings to evaluate the effectiveness of the proposed FedFNN. Collected from multiple scenes, the selected datasets are representative and widely used. The descriptive statistics of each dataset are listed in Table \ref{tbl:dataset_detail}.
In addition, to verify the superior performance of our model, state-of-the-art DFNNs and FL methods for constructing deep models are adopted as comparison methods. The details of all algorithms adopted in this section and their settings are introduced below.
\begin{itemize}
\item[-] DFNN: This is the fully DFNN algorithm proposed in \cite{shi2020consensus}, which adopts consensus learning in both the parameter learning and structure learning procedures and achieves state-of-the-art performance among distributed fuzzy models. As mentioned before, this model learns a consensus FNN for all clients, which limits its applications in non-IID scenarios. We use DFNN+ and DFNN* to denote homogeneous and heterogeneous federated setups, respectively.
\item[-] FedAvg \cite{mcmahan2017communication}: This is the most basic and popular algorithm for developing FL models. Similarly, we use FedAvg+ and FedAvg* to refer to the FedAvg models that cope with homogeneous and heterogeneous scenarios, respectively. Considering that the preferable deep model structures for different datasets vary from each other, to make the comparison convincing, we design more than 20 types of deep models for the comparison and report the best one for each dataset in the following tables and figures.
\item[-] MOON \cite{li2021model}: This is the state-of-the-art FL approach for deep models. MOON adapts individual clients based on their dissimilarity with the server, which bestows the outperforming learning ability on the network model in solving heterogeneous distributed scenarios. Here, we utilize MOON for heterogeneous situations, namely, MOON*, for the comparison. Similar to FedAvg, we choose 20 deep models with different structures and list their best performance, as shown below.
\item[-] FedFNN: This is the model proposed in our paper, which adopts a rule evolution strategy to make full use of each fuzzy rule and acquire better estimation results for each individual dataset. We set the global number of rules $K$ as $15$, the number of ERL iterations as 15, and the number of coevolutionary rounds $L$ as $10$. For the hyperparameter setting, $\bar{\beta}$ is set as $0.7$.
\end{itemize}

In our experiment, all the distributed models are assigned with 5 local clients, each of which can only access their own dataset. To simulate heterogeneous local datasets for these clients, we generate five non-IID local data partitions using the Dirichlet distribution $Dir(\alpha)$, where $\alpha \in (0, 100)$ is the concentration parameter and can be seen as the indicator of the non-IID level of the data. Technically, the sample proportions of all categories for all clients are sampled from $Dir(\alpha)$. The local datasets are thus generated based on random sampling from the original dataset based on the obtained category proportions.

It is worth noting that for the fairness of the experiments, the features in all datasets are first normalized between -1 and 1 using the well-known mapminmax normalization method. Then, a certain proportion of the features is randomly polluted by noise generated from a normal Gaussian distribution to simulate uncertainty in the different datasets. This perturbed sample proportion is considered the uncertainty level and is used to verify the uncertainty processing abilities of the comparison algorithms. In this section, all experiments are conducted with 5-fold cross-validation, and each experiment is repeated 10 times. The final reported results are the mean average precision (mAP) values obtained on the test data during these runs.

\begin{table*}[!ht]
\centering
\caption{Average classification accuracies (\%) and standard deviations achieved by each algorithm on the 7 datasets at a 10\% level of uncertainty}
\begin{tabular}{l|ccccccc}
\hline
\multirow{2}{*}{Algorithm}                                      & \multicolumn{7}{c}{Dataset}                                                                                                                                                                                                                                                                   \\ \cline{2-8}
                                                                & \multicolumn{1}{c|}{GSAD}                & \multicolumn{1}{c|}{SDD}                 & \multicolumn{1}{c|}{SC}                  & \multicolumn{1}{c|}{MGT}                 & \multicolumn{1}{c|}{WFRN}                & \multicolumn{1}{c|}{FM}                  & WIL                         \\ \hline
DFNN+                                                        & \multicolumn{1}{c|}{36.64/1.08}          & \multicolumn{1}{c|}{16.26/0.67}          & \multicolumn{1}{c|}{63.02/4.66}          & \multicolumn{1}{c|}{83.22/0.49}          & \multicolumn{1}{c|}{53.04/1.59}          & \multicolumn{1}{c|}{60.56/6.02}          & 83.40/3.48                  \\ \hline
DFNN*                                                        & \multicolumn{1}{c|}{34.91/3.38}          & \multicolumn{1}{c|}{16.74/1.96}          & \multicolumn{1}{c|}{51.08/7.12}          & \multicolumn{1}{c|}{79.02/3.66}          & \multicolumn{1}{c|}{51.50/4.12}          & \multicolumn{1}{c|}{57.22/5.41}          & 82.20/2.20                  \\ \hline
\multirow{2}{*}{FedFNN*}                                     & \multicolumn{1}{c|}{\textbf{90.13/0.89}} & \multicolumn{1}{c|}{\textbf{75.98/3.49}} & \multicolumn{1}{c|}{\textbf{93.12/4.34}} & \multicolumn{1}{c|}{\textbf{86.58/0.56}} & \multicolumn{1}{c|}{\textbf{91.99/3.56}} & \multicolumn{1}{c|}{\textbf{93.33/2.36}} & \textbf{96.35/0.86}         \\
                                                                & \multicolumn{1}{l|}{({\color{teal}$\uparrow$ $21.57$})}             & \multicolumn{1}{l|}{({\color{teal}$\uparrow$ $2.3$})}             & \multicolumn{1}{l|}{({\color{teal}$\uparrow$ $5.6$})}             & \multicolumn{1}{l|}{({\color{teal}$\uparrow$ $11.28$})}             & \multicolumn{1}{l|}{({\color{teal}$\uparrow$ $13.94$})}             & \multicolumn{1}{l|}{({\color{teal}$\uparrow$ $20.03$})}             & \multicolumn{1}{l}{({\color{teal}$\uparrow$ $5.01$})} \\ \hline
\begin{tabular}[c]{@{}l@{}}MOON*\end{tabular} & \multicolumn{1}{c|}{68.56/0.50}          & \multicolumn{1}{c|}{73.68/3.58}          & \multicolumn{1}{c|}{87.52/4.38}          & \multicolumn{1}{c|}{75.30/3.45}          & \multicolumn{1}{c|}{78.05/1.30}          & \multicolumn{1}{c|}{73.00/2.59}          & 91.34/1.19                  \\ \hline
FedAvg*                                                      & \multicolumn{1}{c|}{65.49/0.62}          & \multicolumn{1}{c|}{63.38/4.02}          & \multicolumn{1}{c|}{79.08/0.35}          & \multicolumn{1}{c|}{64.25/0.56}          & \multicolumn{1}{c|}{71.94/3.16}          & \multicolumn{1}{c|}{63.33/7.07}          & 69.17/5.45                  \\ \hline
FedAvg+                                                      & \multicolumn{1}{c|}{70.80/0.72}          & \multicolumn{1}{c|}{89.43/0.71}          & \multicolumn{1}{c|}{98.35/0.20}          & \multicolumn{1}{c|}{82.14/0.69}          & \multicolumn{1}{c|}{82.69/1.82}          & \multicolumn{1}{c|}{75.67/9.25}          & 94.63/1.41                  \\ \hline
\end{tabular}
\label{tbl:cmpr_nl_1}
\end{table*}

\begin{table*}[!ht]
\centering
\caption{Number of parameters required by each state-of-the-art algorithm}
\begin{tabular}{l|crrrrrr}
\hline
\multirow{2}{*}{Algorithm} & \multicolumn{7}{c}{Dataset}                                                                                                                                                                                            \\ \cline{2-8}
                           & \multicolumn{1}{c|}{GSAD}      & \multicolumn{1}{c|}{SDD}       & \multicolumn{1}{c|}{SC}        & \multicolumn{1}{c|}{MGT}    & \multicolumn{1}{c|}{WFRN}    & \multicolumn{1}{c|}{FM}      & \multicolumn{1}{c}{WIL} \\ \hline
FedDNN                     & \multicolumn{1}{r|}{2,003,462} & \multicolumn{1}{r|}{1,591,051} & \multicolumn{1}{r|}{1,973,255} & \multicolumn{1}{r|}{32,070} & \multicolumn{1}{r|}{539,396} & \multicolumn{1}{r|}{549,124} & 1,578,756               \\ \hline
FedFNN                     & \multicolumn{1}{r|}{15,450}    & \multicolumn{1}{r|}{9,525}     & \multicolumn{1}{r|}{1,320}     & \multicolumn{1}{r|}{630}    & \multicolumn{1}{r|}{2,220}   & \multicolumn{1}{r|}{3,930}   & 690                     \\ \hline
\end{tabular}
\label{tbl:cmpr_n_para}
\end{table*}

\begin{figure}[!htbp]
  \centering
  \includegraphics[width=1\columnwidth]{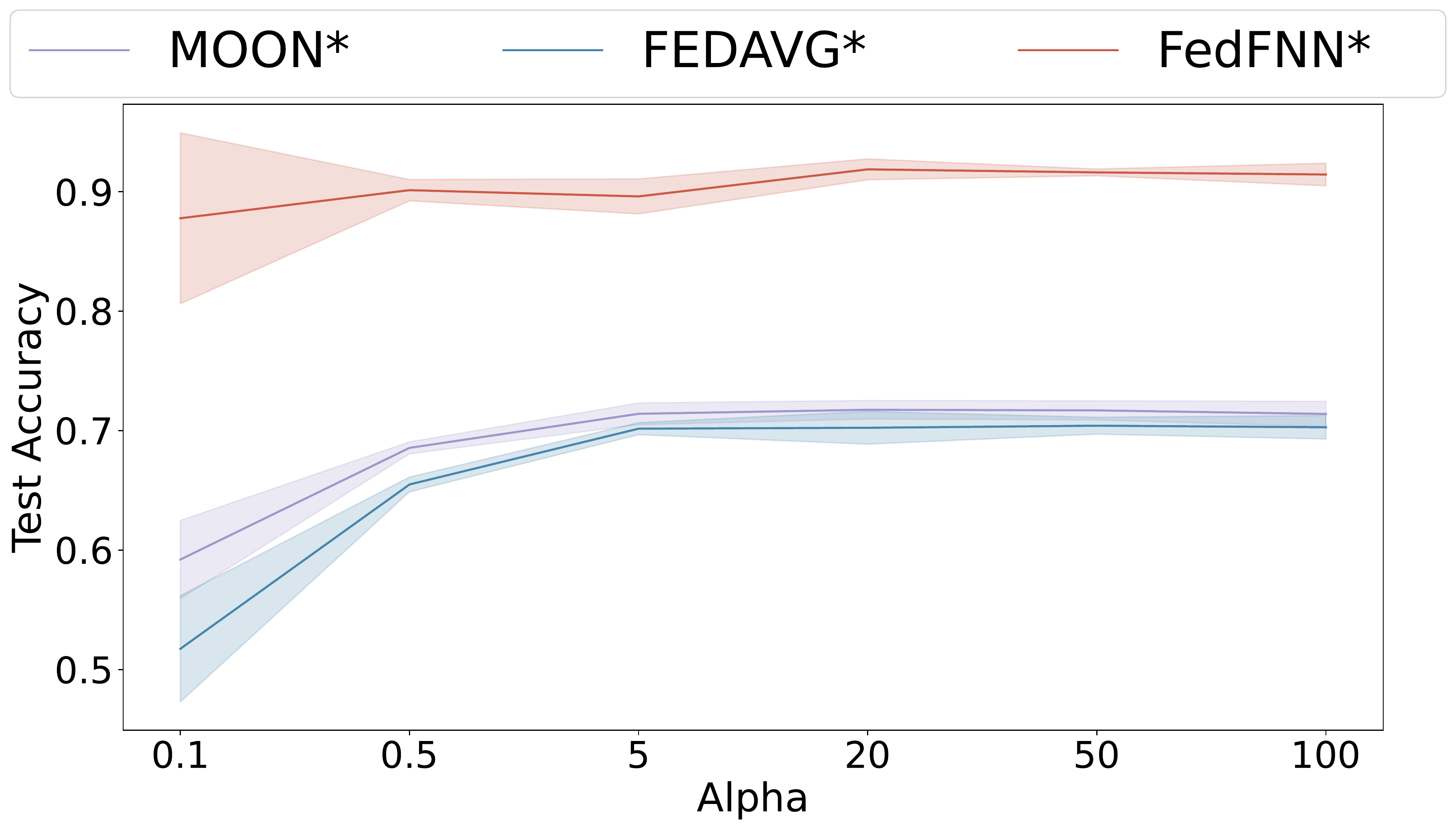}
  \vspace{-10pt}
  \caption{\normalsize Performance of the FedFNN on GSAD at different non-IID levels.}
  \label{fig:hetero_ana}
\end{figure}

\begin{figure}[!htbp]
  \centering
  \includegraphics[width=1\columnwidth]{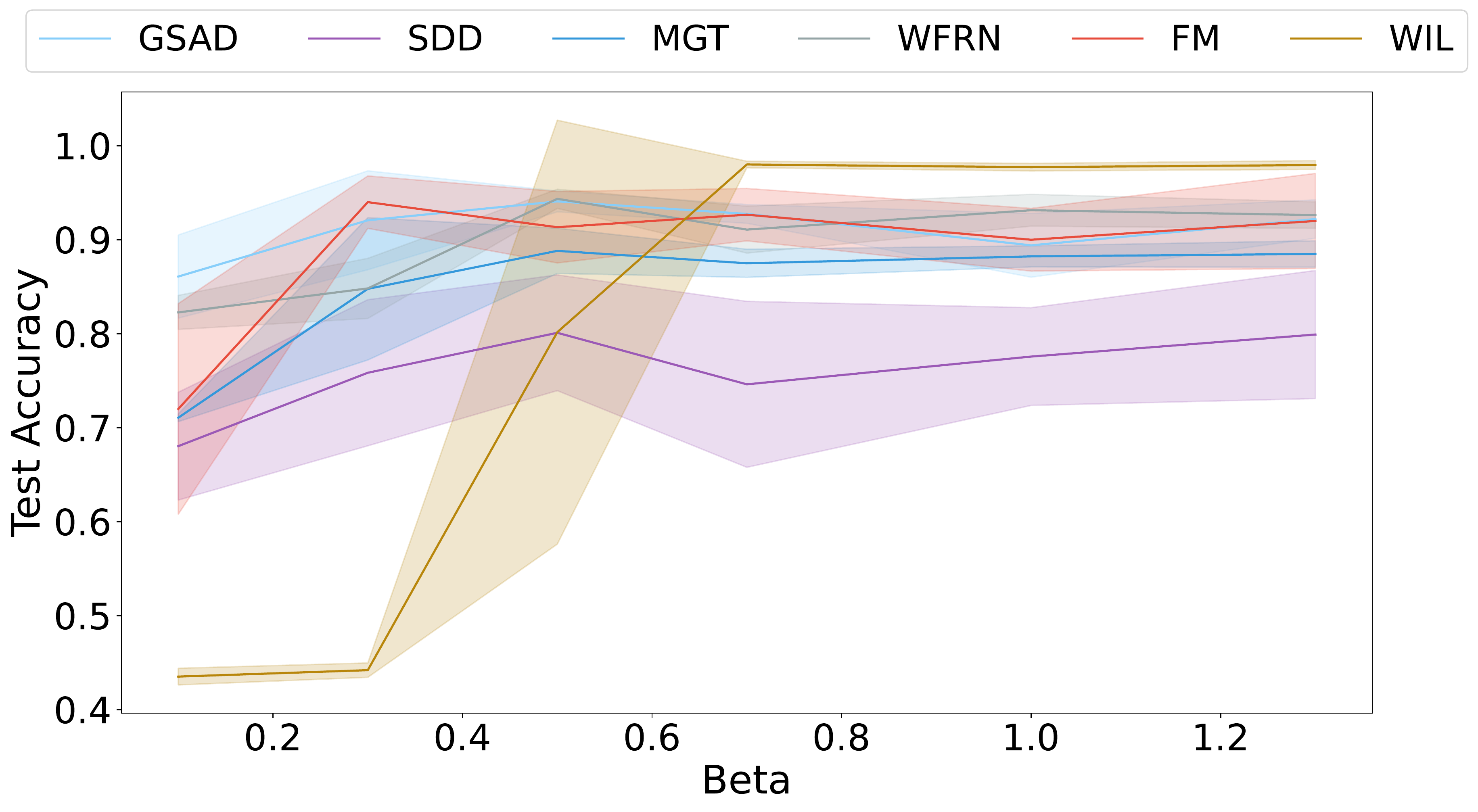}
  \vspace{-10pt}
  \caption{\normalsize Performance of the FedFNN when using different $\beta$.}
  \label{fig:beta_ana}
\end{figure}

\begin{figure}[!htbp]
  \centering
  \includegraphics[width=1\columnwidth]{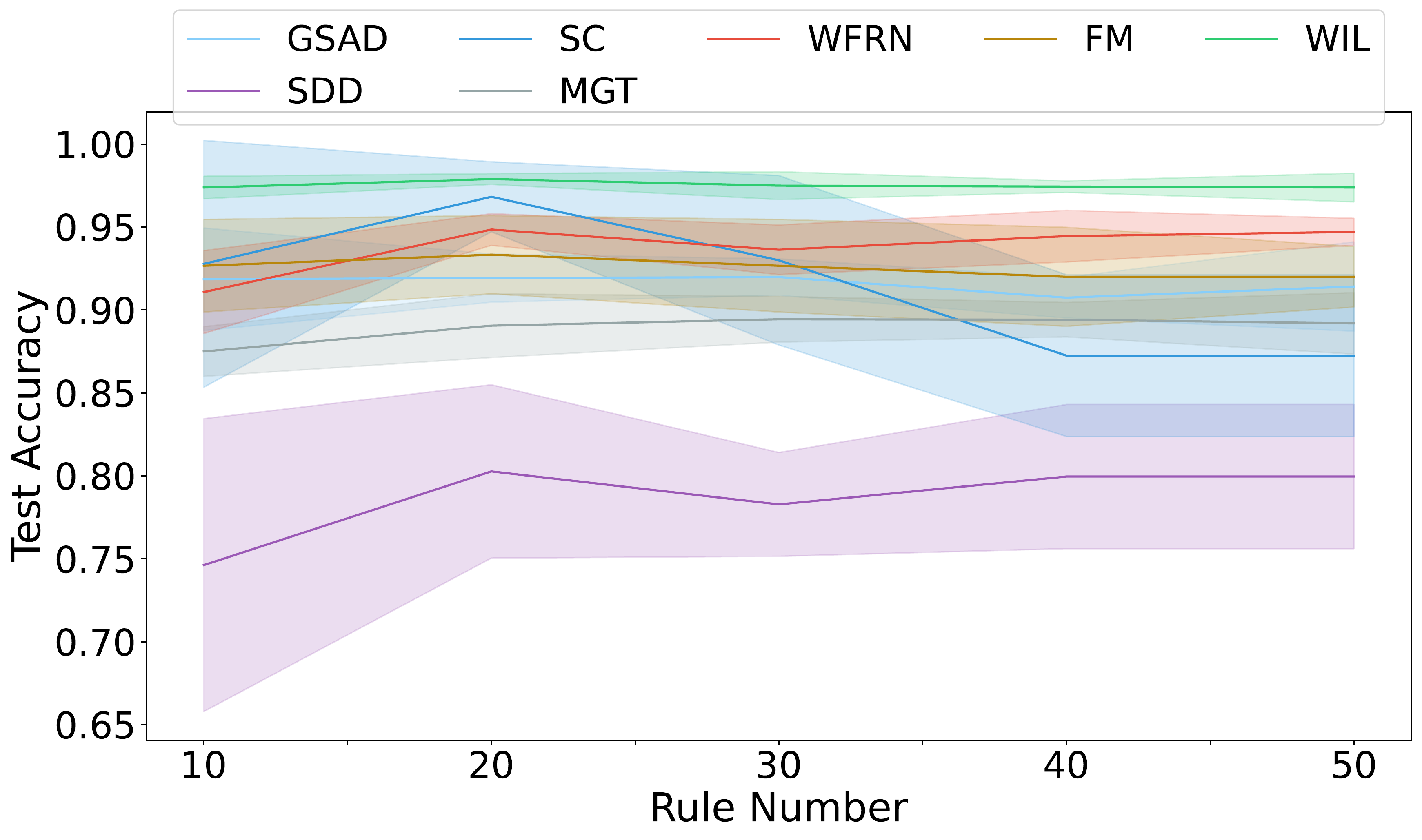}
  \vspace{-10pt}
  \caption{\normalsize Performance of the FedFNN using different initial global rule numbers.}
  \label{fig:n_rule_ana}
\end{figure}

\begin{figure}[!htbp]
  \centering
  \includegraphics[width=1.0\columnwidth]{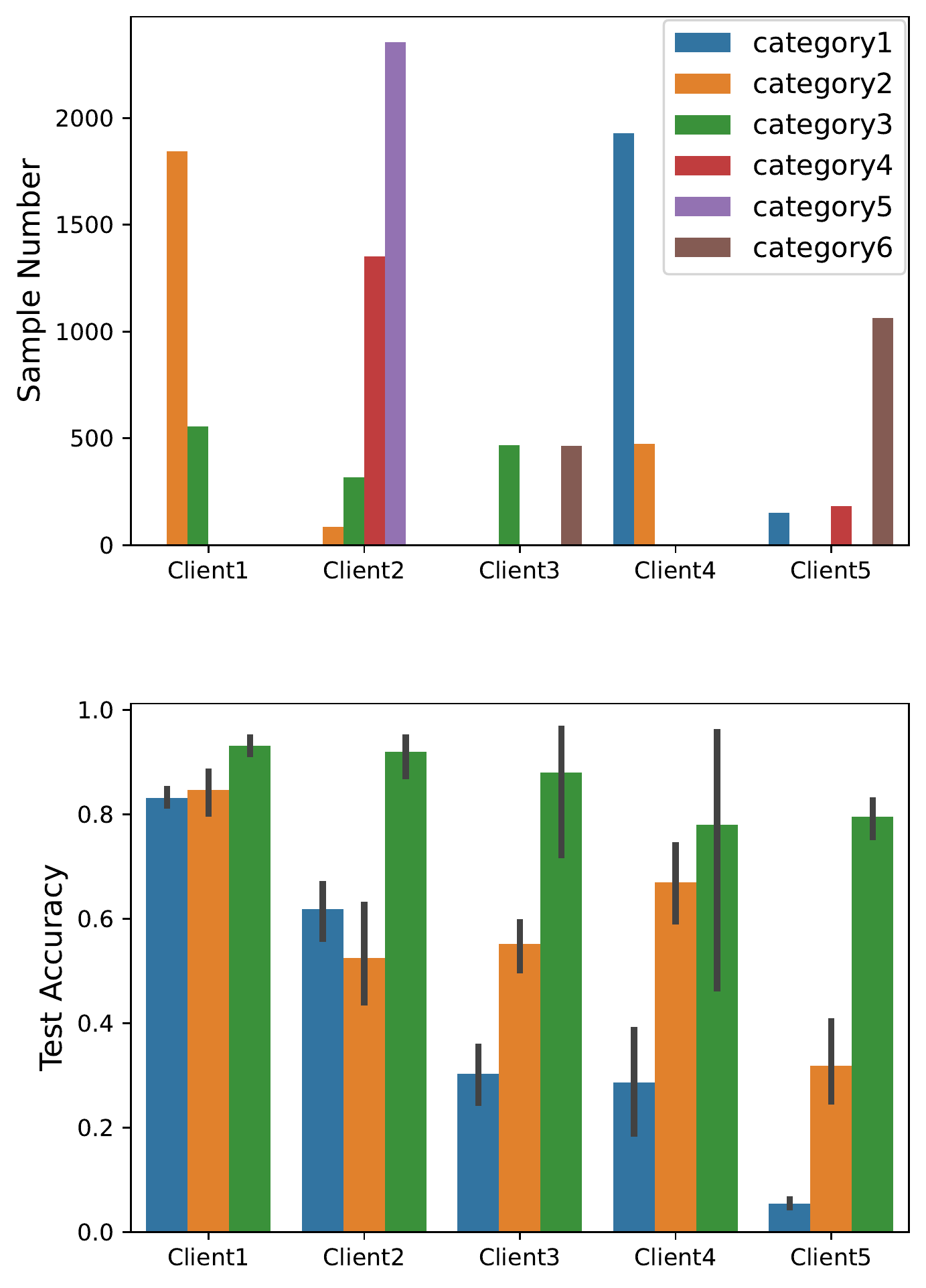}
  \caption{\normalsize Performance of the FedFNN on all clients when training on the GSAD dataset.}
  \label{fig:client_analysis}
\end{figure}

\begin{table}[!ht]
\centering
\caption{Comparison of computing time (s) between FedFNN and DFNN}
\resizebox{\textwidth/2}{!}{
\begin{tabular}{l|cllllll}
\hline
\multirow{2}{*}{Algorithm} & \multicolumn{7}{c}{Dataset}                                                                                                                                                                    \\ \cline{2-8}
                           & \multicolumn{1}{c|}{GSAD} & \multicolumn{1}{c|}{SDD}  & \multicolumn{1}{c|}{SC}   & \multicolumn{1}{c|}{MGT}  & \multicolumn{1}{c|}{WFRN} & \multicolumn{1}{c|}{FM}  & \multicolumn{1}{c}{WIL} \\ \hline
FedFNN                     & \multicolumn{1}{r|}{1805} & \multicolumn{1}{r|}{7259} & \multicolumn{1}{r|}{7050} & \multicolumn{1}{r|}{2348} & \multicolumn{1}{r|}{691}  & \multicolumn{1}{r|}{111} & \multicolumn{1}{r}{260} \\ \hline
DFNN                       & \multicolumn{1}{r|}{683}  & \multicolumn{1}{r|}{1642} & \multicolumn{1}{r|}{1534} & \multicolumn{1}{r|}{545}  & \multicolumn{1}{r|}{178}  & \multicolumn{1}{r|}{38}  & \multicolumn{1}{r}{85}                       \\ \hline
\end{tabular}
}
\label{tbl:time}
\end{table}

\begin{figure*}[!htbp]
  \centering
  \includegraphics[width=2\columnwidth]{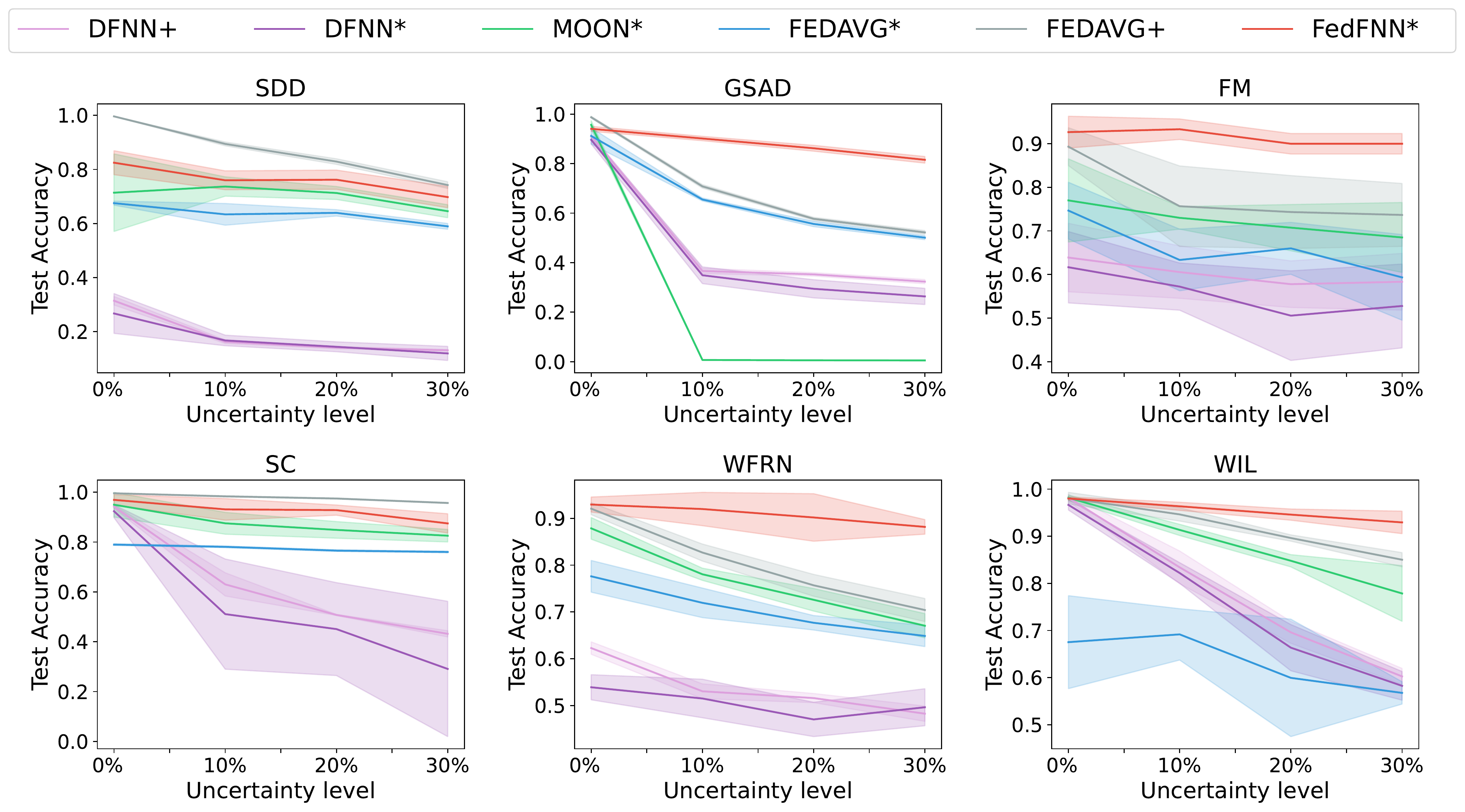}
  \vspace{-10pt}
  \caption{\normalsize Performance of different algorithms when addressing datasets with different uncertainty levels.}
  \label{fig:uncertainty_trend}
\end{figure*}

\begin{figure*}[!htbp]
  \centering
  \includegraphics[width=2\columnwidth]{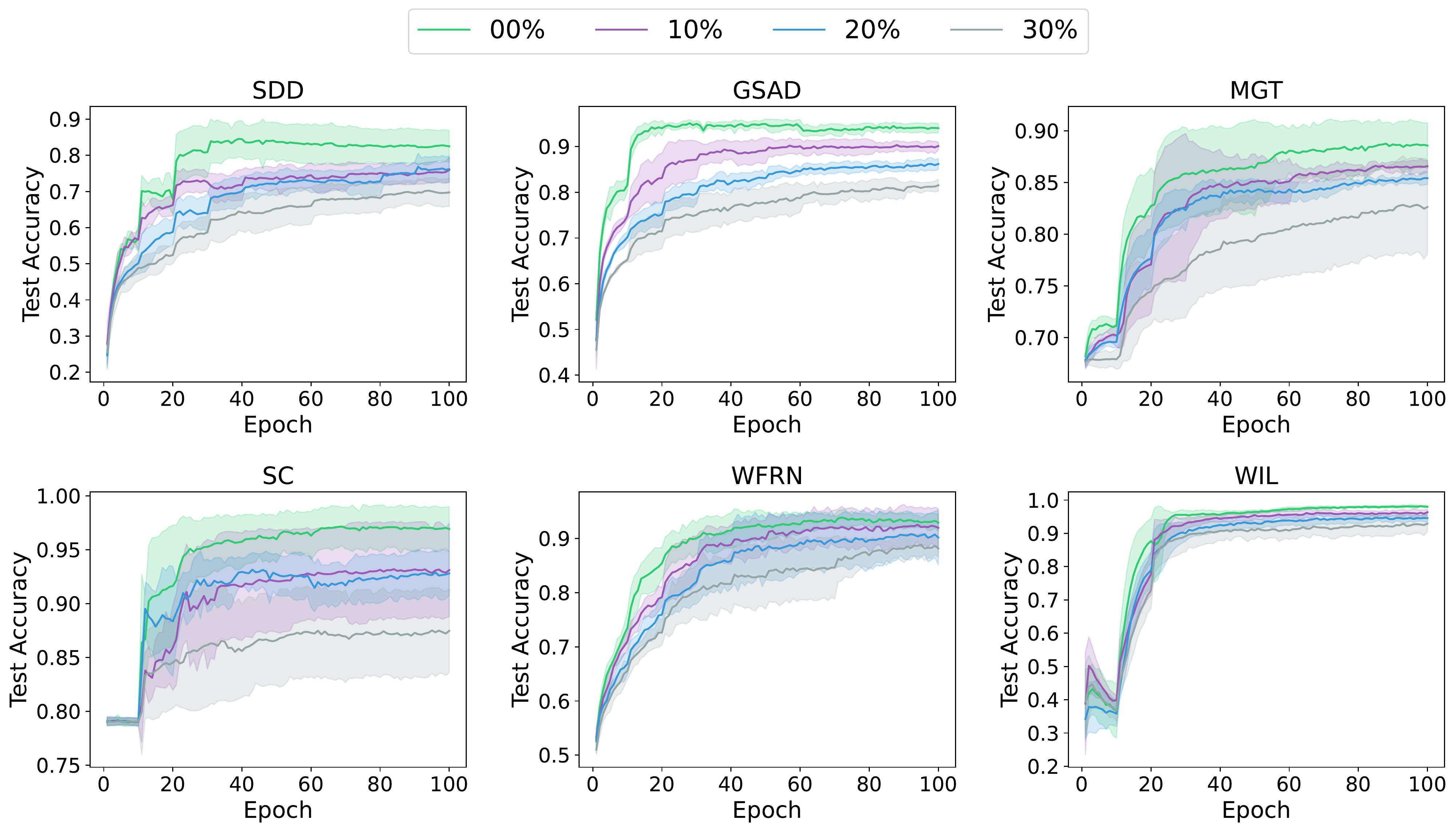}
  \vspace{-10pt}
  \caption{\normalsize The convergence of the FedFNN optimization process under different uncertainty levels on 6 datasets.}
  \label{fig:fedfnn_trend}
\end{figure*}

\subsection{The Performance of the FedFNN on Non-IID Datasets}

We conduct extensive experiments with the aforementioned algorithms on all 7 datasets with a non-IID level of 0.5 and an uncertainty level of 10\% to verify the effectiveness of our FedFNN. The obtained results are summarized in Table \ref{tbl:cmpr_nl_1}, in which the green-colored values are those for which our FedFNN is superior to MOON* on different datasets. According to the table, our model outperforms the existing state-of-the-art FL method MOON* in terms of test accuracy by an average of 11.43\%, which proves the extraordinary heterogeneity handling ability of our FedFNN.

To verify the robustness of our FedFNN when dealing with different levels of non-IID data, we choose the concentration parameter $\alpha$ from the set of $\{0.1,0.5,5,20,50,100\}$ and compare our FedFNN with FedAvg and MOON on the GSAD dataset. The results are shown in Fig. \ref{fig:hetero_ana}, in which the test accuracy curve of the FedFNN for different non-IID levels is flatter and higher than those of the compared FL algorithms. Thus, we can conclude that our FedFNN achieves superior performance when dealing with a wide range of non-IID data. In addition, when the given data are uncertain, our FedFNN* that processes non-IID local data can even beat the FedAvg+ version that processes local IID data on most datasets, which further proves the robustness of our FedFNN.

It is worth noting that our model is far more lightweight than existing FL algorithms. As listed in Table \ref{tbl:cmpr_n_para}, our model has an average of approximately 100 times fewer parameters than existing FL algorithms and achieves a better heterogeneity processing capability.
Besides, we compared the computation overhead of our FedFNN and the DFNN on the Quadro P5000 GPU device with running storage of 26384 MiB. Our results are listed in Table \ref{tbl:time}. Though the ERL consumes extra time compared with DFNN,  our FedFNN increases more than 40\% on average test accuracy than DFNN.

In addition, the personalization abilities of the algorithms are discussed in this section. To explicitly demonstrate the high level of the heterogeneous federated scenario, we plot the number of samples in each category for each local client of the GSAD dataset and the performance achieved by each local model when $\alpha=0.5$ in Fig. \ref{fig:client_analysis}. According to Fig. \ref{fig:client_analysis}, our local models outperform MOON* and FedAvg* on all clients and achieve much higher test accuracy on the 5-th client. MOON* and FedAvg* fail to learn the features of all categories because of the severe data heterogeneity.

Clearly, the global models of MOON and FedAvg tend to deal with the samples in the first five categories and overlook the information learned for the 6th category after several rounds of aggregation; this is because the samples in the 6th category are mostly allocated on client 5, whose learned information is easily disturbed by other clients. Instead, our personalized local FNN for the 5-th client achieves relatively high performance on all clients by automatically generating new rules to estimate the unique samples in the 6-th category. The newly generated rules are unique to this client and cannot be disturbed by other clients; thus, our proposed personalized local FNN exhibits superiority in dealing with heterogeneous local clients based on its flexible structure.

\subsection{The Performance of the FedFNN on Datasets with Uncertainty}

To investigate the effectiveness of the FedFNN when dealing with data uncertainty, we conduct experiments under multiple uncertainty level settings $\{0\%,10\%,20\%,30\%\}$. The performances of all algorithms on all datasets, except the FM dataset, under these uncertainty levels are depicted in Fig. \ref{fig:uncertainty_trend}. Intuitively, our methods achieve state-of-the-art performance for all uncertainty levels and all datasets. In addition, we list the performance attained by all algorithms when processing datasets with 10\% uncertainty in Table \ref{tbl:cmpr_nl_1}. From the table, our model outperforms the DFNN* in terms of test accuracy by an average of 36.40\%. Thus, the extraordinary ability of our FedFNN to deal with data uncertainties can be confirmed.

\subsection{Convergence Analysis of the FedFNN with ERL}

To verify the convergence ability of our proposed FedFNN, the test accuracies it achieves throughout the optimization process are depicted in Fig. \ref{fig:fedfnn_trend}. As the results in this figure show, the FedFNN gradually converges with the increase in the number of communication rounds on all datasets.
In addition, the functions and contributions of the rule cooperation stage and the rule evolution stage are clearly proven in Fig. \ref{fig:fedfnn_trend}. During the $L$-th round of the rule cooperation stage in each ERL iteration, the local clients cooperate with each other to learn more general global rules, and consequently, as shown in Fig. \ref{fig:fedfnn_trend}, the network performance stably improves on all datasets. However, the performance of the local FNNs gradually falls into a bottleneck along with the execution of the rule cooperation stage, as shown in Fig. \ref{fig:fedfnn_trend}, where the test accuracy hardly increases at each ERL iteration. The subsequent rule evolution stage solves this issue by updating the local model architecture. As shown in Fig. \ref{fig:fedfnn_trend}, the performance of the local clients dramatically increases because of the mutation of local rules.
Eventually, these rapid performance improvements induced by the latter learning stage vanish after each local client learns all their required activated local rules.

\subsection{Analysis of Key Parameter Robustness}
We conducted extensive experiments to verify the robustness of our key parameters, including $\alpha$, $\beta$ and the initialized global rule number $K$. Their corresponding results are depicted in Fig. \ref{fig:hetero_ana}, Fig. \ref{fig:beta_ana} and Fig. \ref{fig:n_rule_ana}, respectively. As shown in Fig. \ref{fig:hetero_ana} and Fig. \ref{fig:n_rule_ana}, our FedFNN can still achieve as good performance when setting wide range of parameters. In addition, we can achieve high performance when setting $\beta >= 0.7$ from the results drawn in Fig. \ref{fig:beta_ana}. Intuitively, our FedFNN is proved to be robust to different parameter settings.

\section{Conclusion}
This paper proposes a FedFNN with ERL to handle non-IID issues and data uncertainties in distributed scenarios. The proposed FedFNN integrates fuzzy if-then rules into an FL framework. By considering these fuzzy rules as the basic optimization units, our FedFNN is able to learn a group of general global rules and selectively activate an effective subset of these rules for each local client. This flexible composition approach for fuzzy rules increases the personalization of the local models with respect to handling non-IID data. Inspired by the theory of biological evolution, the proposed ERL method not only encourages the cooperation of local clients at the rule level to improve the generalization of the global rules but also updates the rule activation statuses for all clients to make their local models more personalized. Unlike most existing FL methods that implement aggregation among all local models, the FedFNN with ERL only aggregates the activated rules for their corresponding local clients. This enables the server to selectively aggregate only beneficial local updates, preventing the disturbances brought by harmful updates. Consequently, the FedFNN with ERL provides an effective learning framework for dealing with non-IID issues as well as data uncertainties. Comprehensive experiments verify the superiority and effectiveness of the proposed FedFNN over state-of-the-art methods. In the future, we will improve the ERL to increase the robustness and communication efficiency of FedFNN in scenarios where the agent number is large. In addition, we will investigate the applications in interpretable AI using our FedFNN.

\section{Acknowledgements}
This work was supported in part by the Australian Research Council (ARC) under discovery grant DP210101093 and DP220100803. Research was also sponsored in part by the Australia Defence Innovation Hub under Contract No. P18-650825, Australian Cooperative Research Centres Projects (CRC-P) Round 11 CRCPXI000007, US Office of Naval Research Global under Cooperative Agreement Number ONRG - NICOP - N62909-19-1-2058, and the US Air Force Office of Scientific Research (AFOSR) and Defence Science and Technology (DST) Australian Autonomy Initiative agreement ID10134. We also thank the New South Wales (NSW) Defence Innovation Network and NSW State Government of Australia for financial support in part of this research through grant DINPP2019 S1-03/09 and PP21-22.03.02.

\ifCLASSOPTIONcaptionsoff
  \newpage
\fi

\bibliographystyle{ieeetr}
\bibliography{literatures}

\begin{IEEEbiography}[{\includegraphics[width=1in,height=1.25in,clip,keepaspectratio]{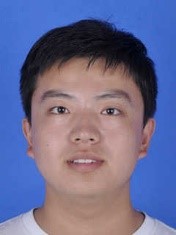}}]
{Leijie Zhang} received the B.S. degree in software engineering from Hebei Normal University, China, in 2015 and the M.S. degree in machine learning from Hangzhou Dianzi University,  China, in 2019. He is currently pursuing the Ph.D. degree in computer science with the University of Technology Sydney, Ultimo, NSW, Australia. His current research interests include fuzzy neural networks and reinforcement learning.
\end{IEEEbiography}

\vspace{-1.5cm}
\begin{IEEEbiography}[{\includegraphics[width=1in,height=1.25in,clip,keepaspectratio]{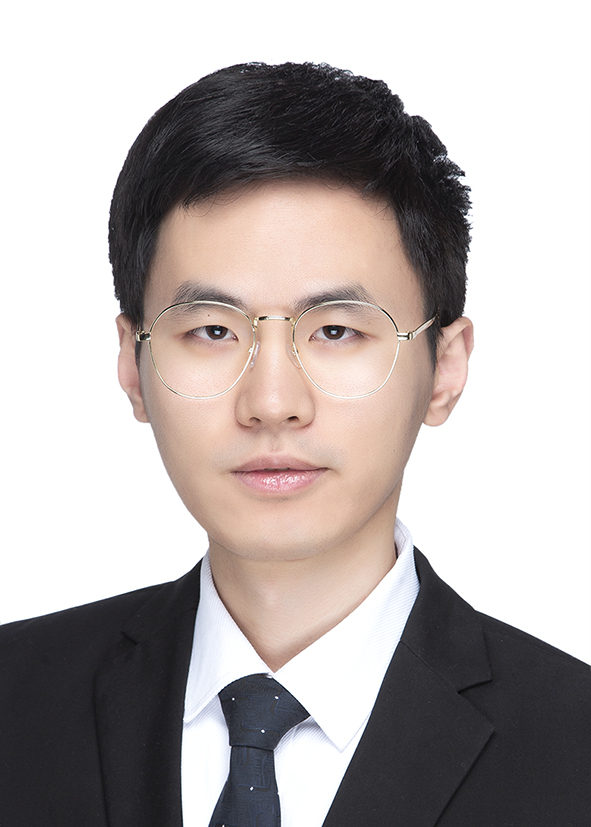}}]
{Ye Shi} received his Ph.D. degree in electrical engineering from the school of Electrical and Data Engineering, University of Technology Sydney (UTS), Australia, 2018. He was a Research Assistant at the University of New South Wales, Australia from 2017 to 2019, and a Postdoctoral Fellow at the University of Technology Sydney from 2019 to 2020. Since January 2021, Dr. Shi has been an Assistant Professor in the School of Information Science and Technology at ShanghaiTech University. His research interests mainly focus on optimization algorithms for Artificial Intelligence, Machine Learning, and Smart Grid. Dr. Shi was a recipient of the Best Paper Award at the 6th IEEE International Conference on Control Systems, Computing, and Engineering in 2016. He serves as a Reviewer for many top-tier journals, such as IEEE TFS, JSAC, TSG, TPS, TII, TIE, and Inf. Sci., Appl. Soft Comput., etc.
\end{IEEEbiography}

\vspace{-1.5cm}
\begin{IEEEbiography}[{\includegraphics*[width=1in, height=1.25in, clip, keepaspectratio]{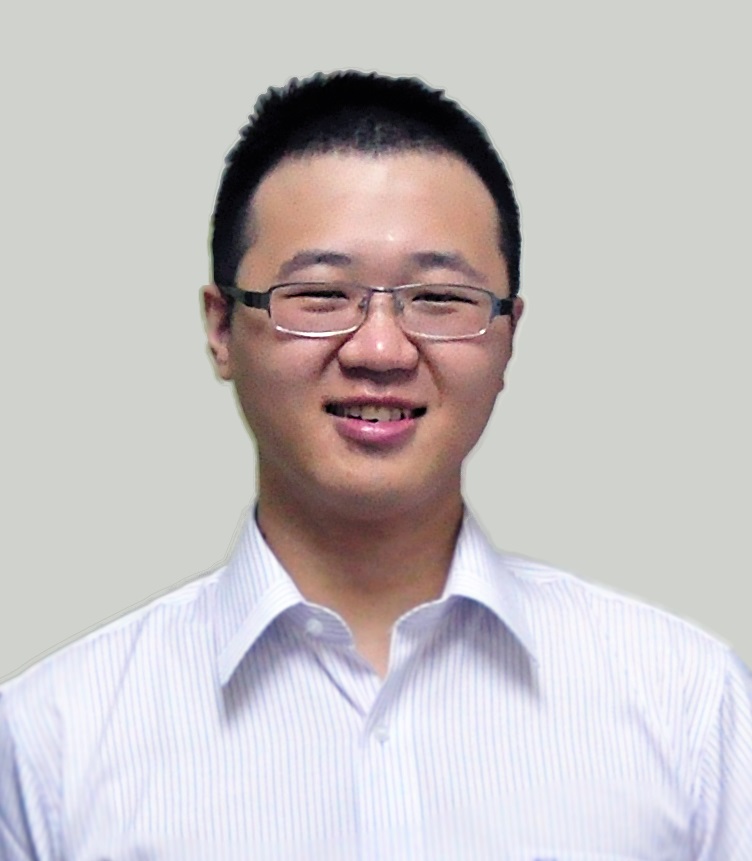}}]
{Yu-Cheng Chang} received the B.S. degree in vehicle engineering from the National Taipei University of Technology, Taipei, Taiwan, in 2008, the M.S. degree with a specialization in system and control from the Department of Electrical Engineering, National Chung-Hsing University, Taiwan, in 2010 and the Ph.D. degree in software engineering from University of Technology Sydney (UTS), Australia in 2021. He currently is a research associate  with the CIBCI lab, UTS. His current research interests include fuzzy systems, human performance modelling and novel human-agent interaction.
\end{IEEEbiography}
\vspace{-1.5cm}
\begin{IEEEbiography}[{\includegraphics[width=1in,height=1.25in,clip,keepaspectratio]{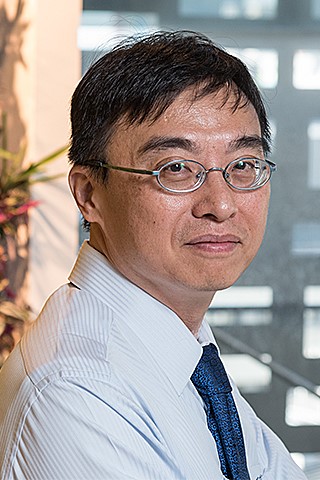}}]
{Chin-Teng Lin} (Distinguished Professor) received a Bachelor’s of Science from National Chiao-Tung University (NCTU), Taiwan in 1986, and holds Master’s and PhD degrees in Electrical Engineering from Purdue University, USA, received in 1989 and 1992, respectively. He is currently a distinguished professor and Co-Director of the Australian Artificial Intelligence Institute within the Faculty of Engineering and Information Technology at the University of Technology Sydney, Australia. He is also an Honorary Chair Professor of Electrical and Computer Engineering at NCTU. For his contributions to biologically inspired information systems, Prof Lin was awarded Fellowship with the IEEE in 2005, and with the International Fuzzy Systems Association (IFSA) in 2012. He received the IEEE Fuzzy Systems Pioneer Award in 2017. He has held notable positions as editor-in-chief of IEEE Transactions on Fuzzy Systems from 2011 to 2016; seats on Board of Governors for the IEEE Circuits and Systems (CAS) Society (2005-2008), IEEE Systems, Man, Cybernetics (SMC) Society (2003-2005), IEEE Computational Intelligence Society (2008-2010); Chair of the IEEE Taipei Section (2009-2010); Chair of IEEE CIS Awards Committee (2022); Distinguished Lecturer with the IEEE CAS Society (2003-2005) and the CIS Society (2015-2017); Chair of the IEEE CIS Distinguished Lecturer Program Committee (2018-2019); Deputy Editor-in-Chief of IEEE Transactions on Circuits and Systems-II (2006-2008); Program Chair of the IEEE International Conference on Systems, Man, and Cybernetics (2005); and General Chair of the 2011 IEEE International Conference on Fuzzy Systems. Prof Lin is the co-author of Neural Fuzzy Systems (Prentice-Hall) and the author of Neural Fuzzy Control Systems with Structure and Parameter Learning (World Scientific). He has published more than 425 journal papers including about 200 IEEE journal papers in the areas of neural networks, fuzzy systems, brain-computer interface, multimedia information processing, cognitive neuro-engineering, and human-machine teaming, that have been cited more than 30,000 times. Currently, his h-index is 83, and his i10-index is 363.

\end{IEEEbiography}
\end{document}